\documentclass[11pt,a4paper]{article}
\usepackage[hyperref]{acl2020}
\usepackage{times}
\usepackage{latexsym}

\usepackage[hang,flushmargin]{footmisc}  
\usepackage{amsfonts,amsmath,amssymb}
\usepackage[T1]{fontenc}  
\usepackage{bold-extra}   
\usepackage{microtype}    
\usepackage{graphicx}
\usepackage{multirow}
\usepackage{bm}           
\usepackage{kotex}        
\usepackage{amsmath}
\usepackage{amssymb}
\usepackage{subcaption}
\usepackage[colorinlistoftodos]{todonotes}

\aclfinalcopy 

\newcommand\LN{\linebreak\noindent}

\title{Analysis of the Penn Korean Universal Dependency Treebank (PKT-UD):\\Manual Revision to Build Robust Parsing Model in Korean}

\author{
  Tae Hwan Oh\textsuperscript{$\spadesuit$}, 
  Ji Yoon Han\textsuperscript{$\spadesuit$}, 
  Hyonsu Choe\textsuperscript{$\spadesuit$}, 
  Seokwon Park\textsuperscript{$\spadesuit$}, 
  Han He,\textsuperscript{$\diamondsuit$} \\
\textbf{\vspace{0.2em}
  Jinho D. Choi\textsuperscript{$\diamondsuit$}, 
  Na-Rae Han\textsuperscript{$\clubsuit$},
  Jena D. Hwang\textsuperscript{$\heartsuit$},
  Hansaem Kim\textsuperscript{$\spadesuit$}} \\
  \textsuperscript{$\spadesuit$}Yonsei University, Seoul, South Korea \\
  \textsuperscript{$\diamondsuit$}Emory University, Atlanta GA 30322, USA \\
  \textsuperscript{$\clubsuit$}University of Pittsburgh, Pittsburgh PA 15260, USA \\
  \textsuperscript{$\heartsuit$}Allen Institute For Artificial Intelligence, Seattle WA 98103, USA \\
  {\small\texttt{\{ghksl0604,clinamen35,choehyonsu,pswon27\}@yonsei.ac.kr}, \texttt{han.he@emory.edu}} \\
  {\small\texttt{jinho.choi@emory.edu}, \texttt{naraehan@pitt.edu}, \texttt{jenah@allenai.org}, \texttt{khss@yonsei.ac.kr}} \\
}

\date{}

\begin{document}
\maketitle

\begin{abstract}

In this paper, we first open on important issues regarding the Penn Korean Universal Treebank (PKT-UD) and address these issues by revising the entire corpus manually with the aim of producing cleaner UD annotations that are more faithful to Korean grammar.
For compatibility to the rest of UD corpora, we follow the UDv2 guidelines, and extensively revise the part-of-speech tags and the dependency relations to reflect morphological features and flexible word-order aspects in Korean.
The original and the revised versions of PKT-UD are experimented with transformer-based parsing models using biaffine attention.
The parsing model trained on the revised corpus shows a significant improvement of 3.0\% in labeled attachment score over the model trained on the previous corpus.
Our error analysis demonstrates that this revision allows the parsing model to learn relations more robustly, reducing several critical errors that used to be made by the previous model.

\end{abstract}


\section{Introduction}
\label{sec:introduction}

In 2018, \citet{chun2018building} published on three dependency treebanks in Korean that followed the latest guidelines from the Universal Dependencies (UD) project, that was UDv2.
These treebanks were automatically derived from the existing treebanks, the Penn Korean Treebank (PKT; \citealt{han2001penn}), the Google UD Treebank \cite{mcdonald2013universal}, and the KAIST Treebank \cite{choi1994kaist}, using head-finding rules and heuristics.

This paper first addresses the known issues in the original Penn Korean UD Treebank, henceforth PKT-UD v2018, through a sampling-based analysis (Section~\ref{sec:corpus}), and then describes the revised guidelines for both part-of-speech tags and dependency relations to handle those issues (Section~\ref{sec:revisions}).
Then, a transformer-based dependency parsing approach using biaffine attention is introduced (Section~\ref{sec:approach}) to experiment on both PKT-UD v2018 and the revised version, henceforth PKT-UD v2020 (Section~\ref{sec:experiments}).
Our analysis shows a significantly reduced number of mispredicted labels by the parsing model developed on PKT-UD v2020 compared to the one developed on PKT-UD v2018, confirming the benefit of this revision in parsing performance.
The contributions of this work are as follows:

\begin{enumerate}
\item Issue checking in PKT-UD v2018.
\item Revised annotation guidelines for Korean and the release of the new corpus, PKT-UD v2020.
\item Development of a robust dependency parsing model using the latest transformer encoder.
\end{enumerate}

\begin{figure*}[htbp!]
\centering

\begin{subfigure}{\textwidth}
\centering
\includegraphics[scale=0.53]{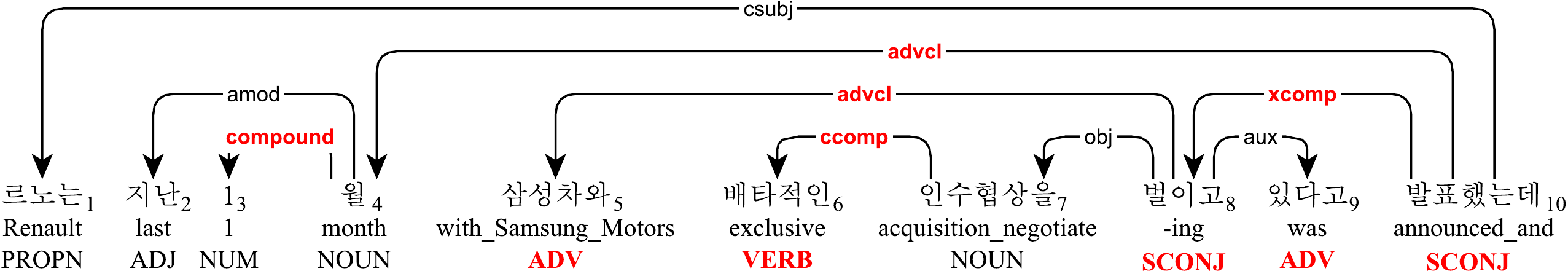}
\caption{Example from v2018 where the labels in revision are indicated by the red bold font.}
\label{sfig:ex-2018a}
\end{subfigure}
\vspace{1ex}

\begin{subfigure}{\textwidth}
\centering
\includegraphics[scale=0.53]{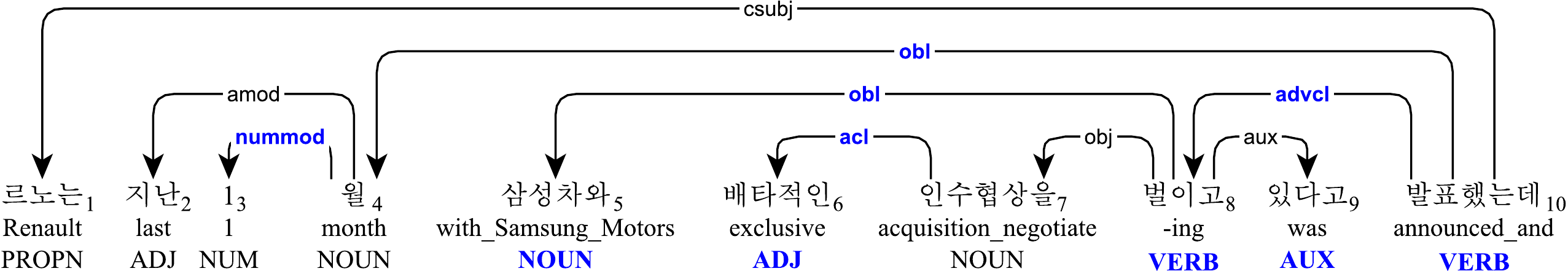}
\caption{Example from v2020 where the revised labels are indicated by the blue bold font.}
\label{sfig:ex-2020a}
\end{subfigure}

\caption{Example from v2018 and v2020, that translates to ``\textit{Renault announced last January that it was negotiating an exclusive acquisition with Samsung Motors, and ...}''. This example continues in Figure~\ref{fig:revision-b}.}
\label{fig:revision-a}
\end{figure*}

\section{Related Works}
\label{sec:related-work}

\subsection{Korean UD Corpora}
\label{ssec:korean-ud-corpora}

According to the UD project website,\footnote{\url{https://universaldependencies.org}} three Korean treebanks are officially registered and released: the Google Korean UD Treebank \cite{mcdonald2013universal}, the Kaist UD Treebank \cite{choi1994kaist}, and the Parallel Universal Dependencies Treebank \cite{zeman-etal-2017-conll}. 
These treebanks were created by converting and modifying the previously existing treebanks. 
The Korean portion of the Google UD Treebank had been re-tokenized into the morpheme level in accordance with other Korean corpora, and systematically corrected for several errors \cite{chun2018building}. 
The Kaist Korean UD Treebank was derived by automatic conversion using head-finding rules and linguistic heuristics \cite{chun2018building}. 
The Parallel Universal Dependencies Treebank was designed for the CoNLL 2017 shared task on Multilingual Parsing, consisting of 1K sentences extracted from newswires and Wikipedia articles.

The Penn Korean UD Treebank and the Sejong UD Treebank were registered on the UD website as well but unreleased due to their license issues.
Similar to the Kaist UD Treebank, the Penn Korean UD Treebank\footnote{The annotation with the word-forms of the Penn Korean UD Treebank can be found here: \url{https://github.com/emorynlp/ud-korean}.} was automatically converted into UD structures from phrase structure trees \cite{chun2018building}.
The Sejong UD Treebank was also automatically converted from the Sejong Corpus, a phrase structure Treebank consisting of 60K sentences from 6 genres \cite{choi2011statistical}. 

\begin{table}[htbp!]
\resizebox{.478\textwidth}{!}{
\begin{tabular}{p{1cm}|p{1cm}|p{1cm}|p{1cm}|p{1cm}|p{1cm}}
\multicolumn{1}{c|}{\textbf{Treebank}} & \multicolumn{1}{c|}{\textbf{GKT}} & \multicolumn{1}{c|}{\textbf{KTB}} & \multicolumn{1}{c|}{\textbf{PUD}} & \multicolumn{1}{c|}{\textbf{PKT}} & \multicolumn{1}{c}{\textbf{Sejong}}  \\
\hline\hline
\multicolumn{1}{c|}{{Sentences}}   &  \multicolumn{1}{c|}{{6k}} &  \multicolumn{1}{c|}{{27k}} & \multicolumn{1}{c|}{{1k}} & \multicolumn{1}{c|}{{5k}} & \multicolumn{1}{c}{{60k}} \\
\hline
\multicolumn{1}{c|}{{Tokens}}   &  \multicolumn{1}{c|}{{80k}} & \multicolumn{1}{c|}{{350k}} & \multicolumn{1}{c|}{{16k}} & \multicolumn{1}{c|}{{132k}} & \multicolumn{1}{c}{{825k}} \\
\hline
\multicolumn{1}{c|}{{Released}}   & \multicolumn{1}{c|}{{O}} &  \multicolumn{1}{c|}{{O}} & \multicolumn{1}{c|}{{O}} & \multicolumn{1}{c|}{{X}} & \multicolumn{1}{c}{{X}} \\
\hline
\multicolumn{1}{c|}{{Unit}}  & \multicolumn{1}{c|}{{Eojeol}} & \multicolumn{1}{c|}{{Eojeol}} & \multicolumn{1}{c|}{{Eojeol}} & \multicolumn{1}{c|}{{Eojeol}} & \multicolumn{1}{c}{{Eojeol}} \\
\hline
\multicolumn{1}{c|}{{Genre}} & Blog, News & Litr, News, Acdm, Mscr & Blog, News & Milt & Litr, News, Acdm, Mscr 
\end{tabular}}
\caption{Korean UD Treebanks. Each abbreviations indicate genres of source texts: webblogs(Blog), newswire(News), literatures(Litr), academic(Acdm), manuscripts(Mscr), military(Milt).}

\label{tbl:koreanUDT}
\vspace{-2ex}
\end{table}

In a related effort, the Electronic and Telecommunication Research Institute (ETRI) in Korea conducted a research on standardizing dependency relations and structures \cite{lim:15a}. 
This effort resulted in the establishment of standard annotation guidelines of Korean dependencies, giving rise to various related efforts that focused on the establishment of Korean UD guidelines that better represent the unique Korean linguistic features.
These studies include \citet{park2018} who focused on the mapping between the UD part-of-speech (POS) tags and the POS tags in the Sejong Treebank, and \citet{Lee2019} and \citet{Oh2019} who provided in-depth discussions of applicability and relevance of UD's dependency relation to Korean.



\subsection{Penn Korean UD Treebank}
As mentioned in Section~\ref{ssec:korean-ud-corpora}, the Penn Korean UD Treebank (PKT-UD v2018) was automatically derived from phrase-structure based the Penn Korean Treebank and the results were published by \citet{chun2018building}. 
Even so, it currently does not number among the Korean UD treebanks officially released corpora under the UD project website.

\noindent Our effort to officially release \citet{chun2018building}'s PKT-UD v2018 has uncovered numerous mechanical errors caused by the automatic conversion and few other unaddressed issues, leading us to a full revision of this corpus. 
PKT-UD v2018 made targeted attempts at addressing a number of language-specific issues regarding complex structures such as empty categories, coordination structures, and allocation of POS tags with respect to dependency relations. However, the efforts were limited, leaving other issues such as handling of copulas, proper allocation of verbs according to their verbal endings, and grammaticalized multi-word expressions were unanswered. 
Thus, this paper aims to address those remaining issues while revising PKT-UD v2018 to clearly represent phenomena in Korean. 


\begin{figure*}[htbp!]
\centering

\begin{subfigure}{\textwidth}
\centering
\includegraphics[scale=0.53]{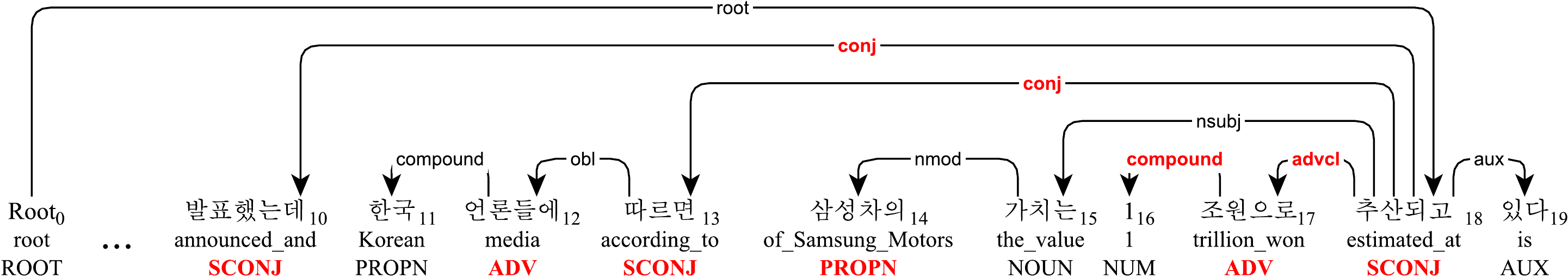}
\caption{Example from v2018 where the labels in revision are indicated by the red bold font.}
\label{sfig:ex-2018b}
\end{subfigure}
\vspace{1ex}

\begin{subfigure}{\textwidth}
\centering
\includegraphics[scale=0.53]{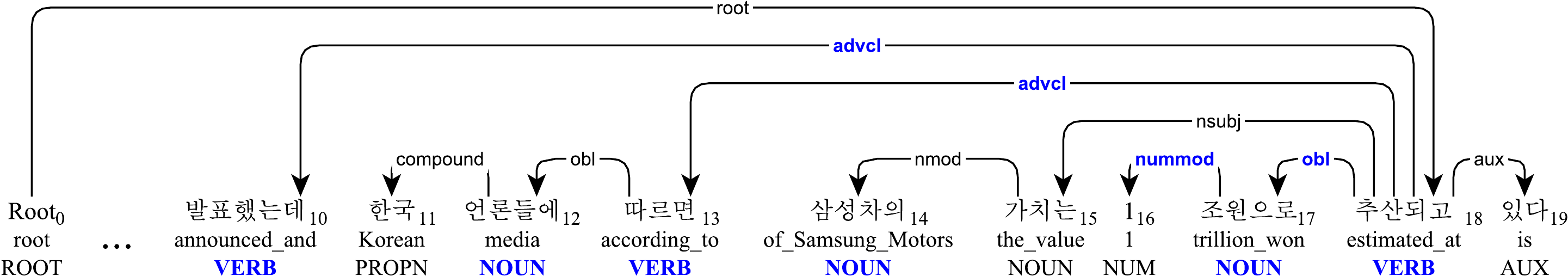}
\caption{Example from v2020 where the revised labels are indicated by the blue bold font.}
\label{sfig:ex-2020b}
\end{subfigure}

\caption{Continuing example from Figure~\ref{fig:revision-a} that translates to ``\textit{... announced ..., and according to Korean media, the value of Samsung Motors is estimated at 1 trillion won}''.}
\label{fig:revision-b}
\end{figure*}


\section{Observations in PKT-UD v2018}
\label{sec:corpus}

The Penn Korean Treebank (PKT) was originally published as a phrase-structure based treebank by \citet{han2001penn}.  
PKT consists of 5,010 sentences from Korean newswire including 132,041 tokens.\footnote{While most Korean resources have what is known as \textit{Eojeol} representing a token and white space is used as delimiter, PTK tokenizes apart symbols, punctuation and even occasional morphemes where strictly required by syntactic structure.}
Following the UDv2 guidelines, \citet{chun2018building} systematically converted PKT to PKT-UD v2018. 
While this effort achieved a measure of success at providing phrase-structure-to-dependency conversion in a manner consistent across three different treebanks with distinct grammatical frameworks, it stopped short of addressing more nuanced issues that arise from aligning grammatical features of Korean, that is a heavily agglutinative language, to the universal standards put forth by UDv2. In building PKT-UD v2018, the POS tags were largely mapped in a categorical manner from the Penn Korean POS tagset. The dependency relations on the other hand were established via head-finding rules that relied on Penn Korean Treebank's existing function tags, phrasal tags, and morphemes.

\citet{chun2018building} did make a few targeted attempts at teasing apart more fine-grained nuances of grammatical functions.
For example, the PKT POS tag (XPOS) \textit{\texttt{DAN}} was subdivided into the UD POS tag (UPOS) \texttt{DET} for demonstrative prenominals (e.g., 이 (\textit{this}), 그 (\textit{the}), and the UPOS \texttt{ADJ} for attributive adjectives (e.g., 새 (\textit{new}), 헌 (\textit{old})) in the recognition that the XPOS \textit{\texttt{DAN}}, focusing primarily on grammatical distribution, conflated two semantically distinct elements. 
However, such efforts were limited in scope, and the project did not examine\LN the full breadth of language-specific issues.

Moreover, the converted annotation was found to contain a share of mechanical errors.  
A case in point, what should have been 5,010 sentences were found to contain 5,036 roots, suggesting low-level parsing errors.
Additionally, a manual examination of the first five sentences in the corpus uncovered a variety of syntactic errors that raised an alarm. 
The worst of the five examined sentences is shown in Figure~\ref{fig:revision-a} (and continued in  Figure~\ref{fig:revision-b}) with errors in both the UPOS and the dependency relation labels (DEPREL). 
While we will not delve into particulars of each error seen in this example, the example provides a general sense for the extent of errors existent that merited our attention.

\noindent These observed issues inspired us to revise PKT-UD v2018, with the aim of producing cleaner syntactic annotations that would be more faithful to the Korean grammar. 
The following section provides specifics of the revision content.

\section{PKT-UD Revisions}
\label{sec:revisions}

\subsection{UPOS Revision}
\label{sec:length}

Revision of the UPOS portion of the resource was done from the ground up.  
That is, instead of correcting PKT-UD v2018's UPOS annotations, we implemented a new mapping from XPOS to UPOS after a careful re-examination of the original mapping schema.
In particular, we consulted the POS mapping guidelines by \citet{park2018} whose morphological tagset, carried over from the Sejong Project \citep{kim2006korean}, differs from PKT's in some key aspects.
However, we found their nuanced view of grammatical characteristics and typology of Korean in reference to the UDv2 very much applicable.  The followings illustrate key ideas of of our UPOS revision approach. 
Below and throughout this paper, we italicize XPOS labels (e.g., \textit{\texttt{DAN}}) so they are visually distinct from UPOS labels (e.g., \texttt{ADJ}).




\paragraph{Copulas mapped to \texttt{ADJ}}

One major target of revision was the scope of the UPOS adjective label \texttt{ADJ} in Korean, which includes typical predicative adjectives such as `예쁘-' (\textit{pretty}) and `다르-' (\textit{different}). 
As mentioned in Section \ref{sec:corpus}, PKT-UD v2018 already extended the \texttt{ADJ} label to include the closed class of adjectives whose distribution is limited to pre-nominal, attributive use which had been grouped together with the determiner category \textit{\texttt{DAN}} in the original PKT. 
In our current work, we further extend the \texttt{ADJ} label to encompass the copula: \textit{\texttt{CO}} (`-이-' (\textit{be})).
In Korean, `-이-' (\textit{be}) is a copula particle that attaches to a nominal to produce a predicate, much like the English `\textit{be}'. 
However, such copula-derived predicates in Korean are known to share semantic and syntactic traits with adjectives rather than verbs, chief among which being their inability to take on the present/habitual aspect verbal ending `-는다` (\textit{do}) which is only allowed on verbs. 
In light of this, we made a decision to map all instances of XPOS' \textit{\texttt{CO}} to UPOS' \texttt{ADJ}.


\paragraph{Consistent \texttt{NOUN} focusing on morpheme roles}

Korean is well-known as an agglutinative language, and \textit{Josa}s (postpositions) are extremely common nominal suffixes that can indicate a variety of syntactic roles of the whole Eojeol unit (Figure \ref{fig:postposition}).
For example, when an adverbial case particle (`에', \textit{\texttt{PAD}}) attaches to a noun, the resulting Eojeol serves the syntactic role of an adverb. When a conjunctive particle (`와', \textit{\texttt{PCJ}}) is used, the Eojeol functions as a noun conjunct. Consequently, PKT-UD v2018 mapped \texttt{ADV} to the former and \texttt{CCONJ} to the latter. 

\vspace{-0.7ex}
\begin{figure}[htp!]
\centering
\begin{tabular}{ccc}
학교 & 학교+\textbf{에} & 학교+\textbf{와}\\
\emph{hakkyo} & \emph{hakkyo+\textbf{PAD}} & \emph{hakkyo+\textbf{PCJ}}\\
(\textit{school}) & (\textit{\textbf{at} school}) & (\textit{school \textbf{and}})\\
\end{tabular}
\caption{\label{fig:postposition}Korean postpositions, marked in bold.} 
\vspace{-0.8ex}
\end{figure}

\noindent However, this distinction underscores a syntactic role rather than a morphological one: while the syntactic role changes with the attachment of the postposition, the POS of the noun itself remains unaffected. UPOS, as a marker that solely demonstrates morphological characteristics of Eojeol rather than its syntactic function, should reflect the morphological status of the nominal. Therefore, we made a decision to allocate the \texttt{NOUN} label to these cases. 

\paragraph{Verbal endings signal \texttt{VERB}}

Korean has verbal endings on predicates that dictate the syntactic role of Eojeol (Figure \ref{fig:verbal-endings}). 
In PKT-UD v2018, predicates marked with \textit{\texttt{ENM}} (nominalization verbal ending) and \textit{\texttt{ECS}} (conjunctive ending) 
are mapped to \texttt{NOUN} and \texttt{SCONJ}, respectively. However, as with the earlier case involving nominals, these verbal ending suffixes should not be treated as fundamentally altering the underlying POS of the predicate itself.
This work revises both cases of UPOS to \texttt{VERB}. Extending the same principle, parallel cases with the same verbal endings involving an adjective or a copula were likewise re-assigned to \texttt{ADJ}.

\begin{figure}[htp!]
\centering
\begin{tabular}{ccc}
먹+\textbf{다} & 먹+\textbf{기} & 먹+\textbf{고}\\
\emph{mek+\textbf{ta}} & \emph{mek+\textbf{ki}} & \emph{mek+\textbf{ko}}\\
(\textit{Eat}) & (\textit{Eat\textbf{ing}}) & (\textit{Eat \textbf{and}})\\
\end{tabular}
\caption{\label{fig:verbal-endings}Korean verbal endings, marked in bold.}
\end{figure}

\paragraph{Statistics of v2018 and v2020}

The complete distributions of PKT-UD v2018 and v2020 are listed in Table~\ref{tbl:upos-stats}.

\begin{table}[htp!]
\centering
\begin{tabular}{c|r|r|r}
\textbf{UPOS} & \multicolumn{1}{c|}{\textbf{v2018}} & \multicolumn{1}{c|}{\textbf{v2020}} & \multicolumn{1}{c}{\textbf{PC}} \\
\hline\hline
\texttt{ADJ}   &  3,431 &  7,034 & 105.0 $\uparrow$ \\
\texttt{ADP}   &  1,251 &  1,425 &  13.9 $\uparrow$ \\
\texttt{ADV}   & 15,174 &  2,851 &  81.2 $\downarrow$ \\
\texttt{AUX}   &  2,263 &  4,060 &  79.4 $\uparrow$ \\
\texttt{CCONJ} &  2,453 &    377 &  84.6 $\downarrow$ \\
\texttt{DET}   &    685 &    685 &   0.0 $\;\:$ \\
\texttt{NOUN}  & 46,866 & 58,367 &  24.5 $\uparrow$ \\
\texttt{NUM}   &  7,931 &  7,602 &   4.1 $\downarrow$ \\
\texttt{PART}  &    464 &    290 &  37.5 $\downarrow$ \\
\texttt{PRON}  &    857 &  1,142 &  33.3 $\uparrow$ \\
\texttt{PROPN} & 12,257 & 12,769 &   4.2 $\uparrow$ \\
\texttt{PUNCT} & 13,428 & 13,428 &   0.0 $\;\:$ \\
\texttt{SCONJ} &  9,780 &    533 &  94.6 $\downarrow$ \\
\texttt{SYM}   &    376 &    376 &   0.0 $\;\:$ \\
\texttt{VERB}  & 13,855 & 21,102 &  52.3 $\uparrow$ \\
\texttt{X}     &    970 &      0 & 100.0 $\downarrow$ \\
\hline
Total & 132,041 & 132,041 & 0.0$\;\;\:$ \\
\end{tabular}
\caption{Universal POS tagset comparison between the 2018 and 2020 versions of the Penn Universal Dependency Treebank. v2018/v2020: the number of tokens in those versions respectively, PC: percentage change.} 
\label{tbl:upos-stats}
\vspace{-2ex}
\end{table}

\subsection{DEPREL Revision}

In re-examining PTK-UD v2018's dependency relations, we consulted two existing dependency annotation guidelines for Korean: \citet{Lee2019} and \citet{Oh2019}. They offer a thorough analysis on applicability of the universal dependency relation labels to Korean, and further identify a list of dependency relations such as \texttt{iobj}, \texttt{xcomp}, \texttt{expl}, and \texttt{cop} (among others) as not suited for capturing characteristics of Korean grammar.
Additionally, where applicable, we took into consideration the UD Japanese Treebank~\citep{asahara2018universal}, since Japanese exhibits many parallel syntactic phenomena as another strictly head-final agglutinative language~\citep{kanayama2018coordinate}.



\paragraph{Reevaluation of \texttt{iobj}} 

We turned our attention to \texttt{iobj}, the DEPREL label for indirect object. We found PKT-UD v2018's decision to assign nominals with dative case markings to \texttt{iobj} questionable, for the following reasons. First, unlike English, where word order distinguishes indirect objects from direct objects (e.g. ``She gave me:\texttt{iobj} a box:\texttt{obj}''), Korean has no such structural constraint that forms the basis for identifying instances of \texttt{iobj}.  The only potential identifier, then, is dative postpositions such as `-에게'(\emph{to}) and `-한테'(\emph{by}), which correspond roughly to English preposition `\textit{to}' as in ``She gave it \textit{\textbf{to}} me''. The problem is, these markers do not exclusively encode the dative case, as seen in examples such as ``개\textbf{에게} 물렸다" (``\textit{I was bit \textbf{by} a dog}''). 

Hence, we adopted a new approach of reassigning all instances of \texttt{iobj} to the oblique relation \texttt{obl}. This move brings language-internal consistency, as postpositions, in many instances, can simply be dropped if contextually recoverable, rendering any such nominals practically indistinguishable from other nominal adverbials that are assigned to \texttt{obl}. This overall approach is also in line with UD Japanese Treebank, where \texttt{iobj} is categorically absent and `に (\textit{ni})', a postposition whose usage largely parallels the two Korean postpositions above, mapping to \texttt{obl}.




\paragraph{Standardizing verbal predicates} 
As shown in Figure \ref{fig:verbal-ending}, 
Korean predicates take on various syntactic functions depending on the attached verbal ending. Predicates with the declarative verbal ending `-다' (\emph{ta}) are assigned to \texttt{root}, which is straightforward. Endings `-은' (\emph{un}) and `-을' (\emph{ul}) on the other hand turn the verb into a modifier to an upcoming noun; the \texttt{acl} relation therefore is the best fit here. Predicates with endings such as `-어서' (\emph{ese}) and `-게' (\emph{key}) modify other predicates, which calls for an \texttt{advcl} assignment.  
In PKT-UD v2018, these cases had received an array of inconsistent allocations such as clausal complements (\texttt{ccomp}/\texttt{xcomp}), auxiliaries (\texttt{aux}), and conjuncts (\texttt{conj}). These were corrected to \texttt{acl} and \texttt{advcl}.


\begin{figure}[htp!]
\centering
\begin{tabular}{ccc}
먹+다 & 먹+은/을 & 먹+어서\\
\emph{mek-ta} & \emph{mek-un/ul} & \emph{mek-ese}\\
(\textit{eat}) & (\textit{ate/to-eat}) & (\textit{eat because})\\
\end{tabular}
\caption{Examples of Korean verbal ending.} 
\label{fig:verbal-ending} 
\end{figure} 

\begin{figure}[htbp!]
\centering
\includegraphics[scale=0.6]{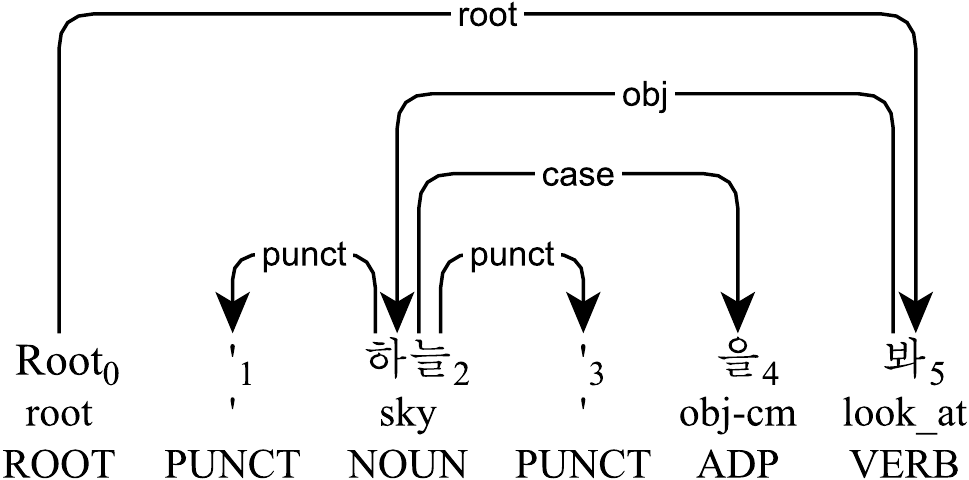}
\caption{PKT-UD v2018 treatment of separated postposition `-을' (\textit{ul}) in "`하늘'을 봐 (Look at the `sky')".} 
\label{fig:orphan}
\end{figure}

\paragraph{Orphaned postpositions and verbal endings} In Korean, verbal endings and postpositions are bound to verbs and nominals, respectively, and cannot occupy their own Eojeol. In natural text, however, they can occasionally be separated from the constituent they attach to via quotation marks, white spaces, or parentheses. PKT-UD v2018 had assigned such orphaned bound morphemes to UPOS of \texttt{PART} (particle) and \texttt{ADP} (adposition) with the DEPREL of \texttt{mark} (marker) and \texttt{case} (case marker), respectively as seen in Figure~\ref{fig:orphan}.


\begin{figure}[htbp!]
\centering
\includegraphics[scale=0.6]{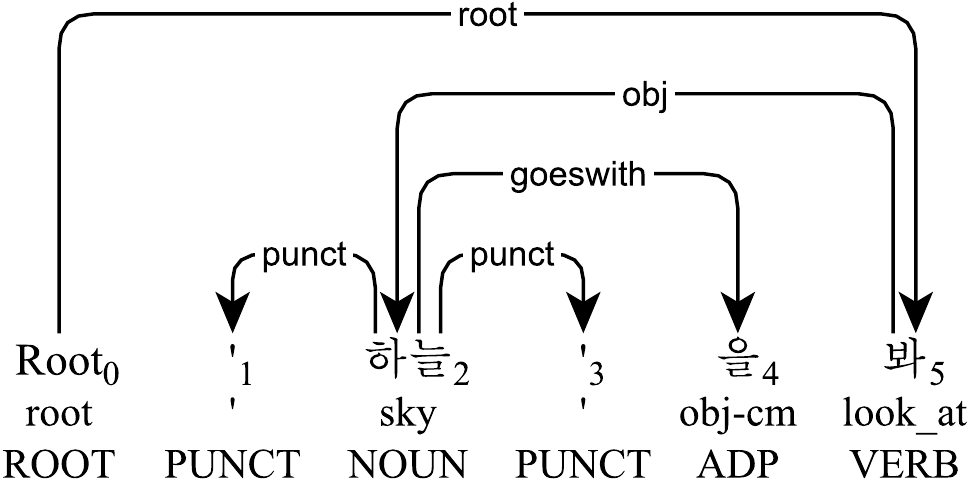}
\caption{Revision of the DEPREL of the separated postposition 을 at "`하늘'을 봐 (Look at the `sky')" in PKT-UD v2020, where \texttt{case} relation for orphaned postposition revised to \texttt{goeswith}.} 
\label{fig:orphan-revised}
\end{figure}

\noindent However, verbal endings and postpositions can express syntactic function only if they are attached to their modifying predicates and nominals. While PKT-UD v2018's assignment of the UPOS and DEPREL are not categorically incorrect, they address morphological relationship between these morphemes rather than their syntactic relationship. That is, even if these bound morphemes are notationally distanced from their heads by punctuation or white spaces, they form a single syntactic unit with their nominals and postpositions. Hence, \texttt{mark} and \texttt{case} were updated to \texttt{goeswith}, used for divided words as seen in Figure~\ref{fig:orphan-revised}, making it clear that the seemingly separate Eojeols (e.g. nominal  and postposition) are actually one unit.


\paragraph{Orphaned copulas}
Similar revisions were applied to copulas. Korean copula morpheme `-이-' (\textit{i}) combines with a nominal on the left and a verbal ending to the right. These copulas too can occasionally be detached via intervening punctuation or white space. To such cases, PKT-UD v2018 had assigned \texttt{cop} as the DEPREL. These instances have been updated to \texttt{goeswith} in accordance with the treatment given to verbal endings and postpostions.


\paragraph{\texttt{root}s and \texttt{flat}s}
The number of \texttt{root} is adjusted from 5,036 to 5,010 after correcting sentences with zero or more roots. Additionally, DEPREL of Eojeols that used to be incorrectly mapped to \texttt{compound} are now assigned to \texttt{flat}. 

\paragraph{Statistics of v2018 and v2020}
The complete DEPREL distributions of PKT-UD v2018 and v2020 are listed in Table~\ref{tbl:udep-stats}.

\begin{table}[htp!]
\centering
\begin{tabular}{c|r|r|r}
\textbf{DEPREL} & \multicolumn{1}{c|}{\textbf{v2018}} & \multicolumn{1}{c|}{\textbf{v2020}} & \multicolumn{1}{c}{\textbf{PC}} \\
\hline\hline
\texttt{acl}      &  1,488 & 11,210 &    653.4 $\uparrow$ \\
\texttt{advcl}    & 11,636 &  5,086 &     56.3 $\downarrow$ \\
\texttt{advmod}   &  2,964 &  3,125 &     5.4 $\uparrow$ \\
\texttt{amod}     &  1,595 &  1,593 &     0.1 $\downarrow$ \\
\texttt{appos}    &  1,182 &  1,173 &     0.8 $\downarrow$ \\
\texttt{aux}      &  4,807 &  4,061 &    15.5 $\downarrow$ \\
\texttt{case}     &  1,548 &      0 &   100.0 $\downarrow$ \\
\texttt{ccomp}    &  9,858 &  1,989 &    79.8 $\downarrow$ \\
\texttt{cc}       &    785 &    473 &    39.7 $\downarrow$ \\
\texttt{compound} & 28,908 & 21,433 &    25.9 $\downarrow$ \\
\texttt{conj}     &  9,960 &  7,155 &    28.2 $\downarrow$ \\
\texttt{cop}      &    418 &      0 &   100.0 $\downarrow$ \\
\texttt{csubj}    &  8,014 &  8,012 &     0.0 $\downarrow$ \\
\texttt{dep}      &    609 &     10 &    98.4 $\downarrow$ \\
\texttt{det}      &    685 &    685 &     0.0 $\;\:$ \\
\texttt{fixed}    &    528 &    589 &    11.6 $\uparrow$ \\
\texttt{flat}     &     18 &    739 & 4,005.6 $\uparrow$ \\
\texttt{goeswith} &      0 &  2,199 &   100.0 $\uparrow$ \\
\texttt{iobj}     &    222 &      0 &   100.0 $\downarrow$ \\
\texttt{mark}     &  1,003 &      0 &   100.0 $\downarrow$ \\
\texttt{nmod}     &  5,555 &  5,501 &     1.0 $\downarrow$ \\
\texttt{nsubj}    &  4,012 &  4,114 &     2.5 $\uparrow$ \\
\texttt{nummod}   &    154 &  7,341 & 4,666.9 $\uparrow$ \\
\texttt{obj}      &  9,823 &  9,849 &     0.3 $\uparrow$ \\
\texttt{obl}      &  3,357 & 16,891 &   403.2 $\uparrow$ \\
\texttt{orphan}   &     0 &       9 &   100.0 $\uparrow$ \\
\texttt{punct}    & 13,073 & 13,794 &     5.5 $\uparrow$ \\
\texttt{root}     &  5,036 &  5,010 &     0.5 $\downarrow$ \\
\texttt{xcomp}    &  4,803 &      0 &   100.0 $\downarrow$ \\
\hline
Total & 132,041 & 132,041 & 0.0 $\;\:$ \\
\end{tabular}
\caption{Universal dependency label comparison between v2018 and v2020 of the Penn Universal Dependency Treebank. v2018/v2020: the number of tokens in those versions respectively, PC: percentage change.} 
\label{tbl:udep-stats}
\end{table}

\section{Parsing Approach}
\label{sec:approach}

\begin{figure*}[htbp!]
\centering
\includegraphics[width=\textwidth]{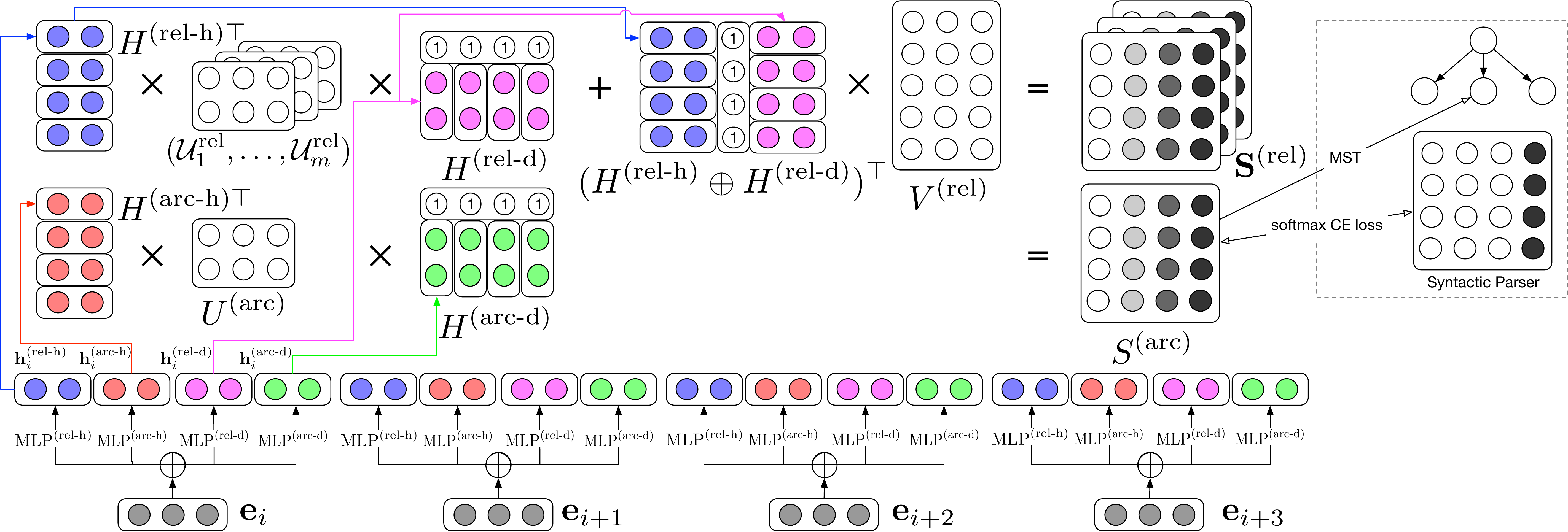}
\caption{The overview of our transformer-based biaffine dependency parsing model.}
\label{fig:biaffine}
\end{figure*}

Our dependency parsing model is based on the biaffine parser using contextualized embeddings such as BERT \cite{devlin-etal-2019-bert} that has shown the state-of-the-art results on both syntactic and semantic dependency parsing tasks in multiple languages \cite{he-choi-2019}.
This model is simplified from the original biaffine parser introduced by \citet{dozat:17a} such that trainable token embeddings are removed and lemmas are used instead of word forms.
This section proposes an even more simplified model that no longer uses embeddings from POS tags, so it can be easily adapted to languages that do not have dedicated POS taggers, and drops the Bidirectional LSTM encoder while integrating the transformer layers directly into the biaffine decoder so that it minimizes the redundancy of having multiple encoders for the generation of contextualized embeddings.

Given an input sentence, every token $w_i$ is first segmented into one or more sub-tokens by the SentencePiece tokenizer \cite{kudo-richardson-2018-sentencepiece} and fed into a transformer.
The output embedding corresponding to the first sub-token of $w_i$ is treated as the embedding representation of $w_i$, say $\mathbf{e}_i$, and fed into four types of multilayer perceptron (MLP) layers to extract features for $w_i$ being a head $\text{(*-h)}$ or a dependent $\text{(*-d)}$ for the arc relations ($\text{arc-*}$) and the labels ($\text{rel-*}$) ($k$ and $l$ are the dimensions of the arc and label representations, respectively):

\begin{align*}
\mathbf{h}_i^\text{(arc-h)} &= \mathrm{MLP}^\text{(arc-h)}(\mathbf{e}_i) \in \mathbb{R}^{k \times 1}\\
\mathbf{h}_i^\text{(arc-d)} &= \mathrm{MLP}^\text{(arc-d)}(\mathbf{e}_i) \in \mathbb{R}^{k \times 1}\\
\mathbf{h}_i^\text{(rel-h)} &= \mathrm{MLP}^\text{(rel-h)}(\mathbf{e}_i) \in \mathbb{R}^{l \times 1}\\
\mathbf{h}_i^\text{(rel-d)} &= \mathrm{MLP}^\text{(rel-d)}(\mathbf{e}_i) \in \mathbb{R}^{l \times 1}
\end{align*}

\noindent All feature vectors, $\mathbf{h}_1^\text{*}, \ldots, \mathbf{h}_n^\text{*}$, from each representation are stacked into a matrix ($n$ is the number of tokens in a sentence); these matrices together are used to predict dependency relations among every token pairs.
Note that bias terms are appended to the feature vectors $\mathbf{h}_i^\text{(*-d)}$ that represent dependent nodes to estimate the likelihood of a certain relation given only the head node:

\begin{align*}
H^\text{(arc-h)} &= (\mathbf{h}_1^\text{(arc-h)}, \ldots, \mathbf{h}_n^\text{(arc-h)}) \in \mathbb{R}^{k \times n} \\
H^\text{(arc-d)} &= (\mathbf{h}_1^\text{(arc-d)}, \ldots, \mathbf{h}_n^\text{(arc-d)}) \oplus \mathbf{1} \in \mathbb{R}^{(k+1) \times n} \\
H^\text{(rel-h)} &= (\mathbf{h}_1^\text{(rel-h)}, \ldots, \mathbf{h}_n^\text{(rel-h)}) \in \mathbb{R}^{l \times n} \\
H^\text{(rel-d)} &= (\mathbf{h}_1^\text{(rel-d)}, \ldots, \mathbf{h}_n^\text{(rel-d)}) \oplus \mathbf{1} \in \mathbb{R}^{(l+1) \times n} \\
\end{align*}

\noindent The bilinear and biaffine classifiers are then used for the arc and label predictions respectively, where $U^{(\text{arc})}$, $U_i^\text{(rel)}$ and $V^{(\text{rel})}$ are trainable parameters, and $m$ is the number of dependency labels. 
In particular, a separate weight matrix $U_i^\text{(rel)}$ is dedicated for the prediction of each label: 

\begin{align*}
S^\text{(arc)} &= H^{\text{(arc-h)}\top} \cdot U^{(\text{arc})} \cdot H^{\text{(arc-d)}} \in \mathbb{R}^{n \times n}  \\
\mathcal{U}_i^\text{(rel)} &= H^{(\text{rel-h})\top} \cdot U_i^\text{(rel)} \cdot H^{(\text{rel-d})} \in \mathbb{R}^{n \times n} \\
\mathbf{S}^{\text{(rel)}} &= (\mathcal{U}_1^\text{(rel)}, \ldots, \mathcal{U}_m^\text{(rel)})\\
                 &+ (H^{(\text{rel-h})} \oplus H^{(\text{rel-d})})^\top \cdot V^{(\text{rel})} \in \mathbb{R}^{m \times n \times n}
\end{align*}

\noindent Once the arc score matrix $S^\text{(arc)}$ and the label score tensor $\mathbf{S}^{\text{(rel)}}$ are generated by those classifiers, the Chu-Liu-Edmond's maximum spanning tree (MST) algorithm is applied to $S^\text{(arc)}$ for the arc prediction, then the label with largest score in $\mathbf{S}^{\text{(rel)}}$ corresponding to the arc is taken for the label prediction:

\begin{align*}
  arc &= \text{MST}(S^\text{(arc)})  \\
label &= \text{argmax}(\mathbf{S}^{\text{(rel)}}[\text{index}(arc)])
\end{align*}

\section{Experiments}
\label{sec:experiments}

To extrinsically assess the quality of our revision, parsing models are separately developed on PKT-UD v2018 and v2020; in other words, v2018 models are trained and evaluated on PKT-UD v2018 whereas v2020 models are trained and evaluated on PKT-UD v2020.
The transformer-based parsing approach in Section~\ref{sec:approach} is used to develop all models.
For each version of the corpus, three models are developed by initializing neural weights with different random seeds and the average accuracy and its standard deviation is reported for each version.
The entire corpus is divided into the training (\texttt{TRN}), development (\texttt{DEV}), and evaluation  (\texttt{TST}) sets by following the 80/10/10\% split (Table~\ref{tbl:data-stats}).

\begin{table}[htbp!]
\centering\small 
\begin{tabular}{l||r|r|r}
 & \multicolumn{1}{c|}{\bf\texttt{TRN}} & \multicolumn{1}{c|}{\bf\texttt{DEV}} & \multicolumn{1}{c}{\bf\texttt{TST}} \\
\hline\hline
\# of Sentences &   4,010 &    501 &    500 \\
\# of Tokens    & 105,947 & 13,088 & 13,023 \\
\end{tabular}
\caption{Statistics of the data split.}
\label{tbl:data-stats}
\end{table}

\noindent The multilingual BERT\footnote{ \url{https://github.com/google-research/bert/blob/master/multilingual.md}} is used as the transformer encoder in our parsing models \cite{devlin-etal-2019-bert}.
All models are optimized by the sum of softmax cross-entropy losses on the gold dependency heads and labels.
AdamW \cite{loshchilov2018decoupled} is used as the optimizer with the learning rate of 5e-06 for the BERT weights and 5e-05 for the rest. 
The learning rate is scheduled as a combination of both linear warm-up and decay phases. 
The models are trained for 100 epochs with a batch size of 150.
Following the standard practice, we evaluate our best models with the unicode punctuation ignored using the unlabeled attachment score (UAS) and the labeled attachment score (LAS).

\noindent Table~\ref{tbl:results} shows the results achieved by the v2018 and v2020 models.
The v2020 model shows a significantly improvement of 3.0\% in LAS over the v2018 model.
This makes sense because the major parts of the revision are dedicated to DEPREL consistency, yielding more robust parsing performance in labeling.
The v2020 model also gives a good improvement of 0.6\% in finding dependency arcs.
The improved parsing results ensure the higher quality annotation in PKT-UD v2020 that is  encouraging.

\begin{table}[htbp!]
\centering 
\begin{tabular}{l||c|c}
\multicolumn{1}{c|}{\bf } & \multicolumn{1}{c|}{\bf UAS} & \multicolumn{1}{c}{\bf LAS} \\
\hline\hline
v2018 & 90.7 ($\pm$0.2) & 86.0 ($\pm$0.1) \\
v2020 & \textbf{91.3} ($\pm$0.1) & \textbf{89.0} ($\pm$0.1) \\
\end{tabular}
\caption{Results by the v2018 and v2020 models.}
\label{tbl:results}
\end{table}
\section{Error analysis}
\label{sec:Error analysis}

\paragraph{PKT-UD v2018}

We perform an error analysis on the parsing outputs generated by the v2018 model.
Our analysis shows that the head error occurred in 1,360 Eojeols and the label error occurred in 4,292 Eojeols. 
Table~\ref{tbl:deprel-error} shows the distribution of head and label errors per label based on the revised test set.
The relations \texttt{advcl}, \texttt{nummod}, \texttt{acl}, and \texttt{obl} have a high error rate, which are due to the inconsistencies seen in the data we handled by establishing clear criteria. Moreover, the labels \texttt{goeswith} and \texttt{flat} saw 100\% error, again, due to the errors we observed during the revision process.

\begin{table}[htp!]
\centering
\begin{tabular}{c|r|r}
\textbf{DEPREL} & \multicolumn{1}{c|}{\textbf{Error}} & {\textbf{Percentage}}  \\
\hline\hline
\texttt{obl}   &  1,294 &  30.15\% \\
\texttt{acl}   &  961 &  22.39\% \\
\texttt{nummod}   & 777 &  18.1\% \\
\texttt{advcl}   &  462 &  10.76\% \\
\texttt{goeswith} &  203 &  4.73\% \\
\texttt{conj}   &    99 &  2.31\% \\
\texttt{compound}  & 96 & 2.24\% \\
\texttt{flat}   &  91 & 2.12\% \\
\texttt{ccomp}  &    77 & 1.79\% \\
\texttt{etc}  &    232 & 5.41\% \\
\hline
Total & 4,292 & 100\% \\
\end{tabular}
\caption{DEPREL error of PKT-UD v2018.} 
\label{tbl:deprel-error}
\vspace{-2ex}
\end{table}


\noindent There is an observable trend in these errors.
For example, a number of error cases report \texttt{advcl} as \texttt{xcomp}, \texttt{conj}, or \texttt{ccomp} while \texttt{nummod} tends to be wrongly parsed to \texttt{compound}, \texttt{acl} to \texttt{ccomp}, and \texttt{obl} to \texttt{advcl}. Multiple cases of parsing errors due to errors in the UPOS are also found. Incorrect UPOS appears to commit errors while allocating edge and DEPREL. The annotation guideline based on XPOS is already described  in Section~\ref{sec:revisions}.

\paragraph{PKT-UD v2020}
After revising the data according to the criteria presented in Section~\ref{sec:revisions}, many improvements have been made. The error rate of \texttt{advcl} decreased from 98.93\% to 2.36\%, the \texttt{nummod} also decreased significantly from 97.37\% to 0.5\%, and the \texttt{acl} error from 86.73\% to 0.9\%. The error rate of \texttt{obl} was also reduced from 79.14\% to 5.5\%. In addition, the error rate is reduced for \texttt{goeswith} and \texttt{flat}. In the case of \texttt{ccomp}, errors decreased by more than 35\% from 44.51\% to 8.67\%. 
These results is indicative of the effect of improving training data by ensuring consistency of annotations.

\begin{table}[htp!]
\centering
\begin{tabular}{c|r|r}
\textbf{DEPREL} & \multicolumn{1}{c|}{\textbf{v2018}} & {\textbf{v2020}}  \\
\hline\hline
\texttt{obl}   &  1,294 &  90 \\
\texttt{acl}   &  961 &  10 \\
\texttt{nummod}   & 777 &  4 \\
\texttt{advcl}   &  462 &  11 \\
\texttt{goeswith} &  203 &  3 \\
\texttt{conj}   &    99 & 85 \\
\texttt{compound}  & 96 & 83 \\
\texttt{flat}   &  91 & 34 \\
\texttt{ccomp}  &    77 & 15 \\
\texttt{etc}  &    232 & 134 \\
\hline
Total & 4,292 & 469 \\
\end{tabular}
\caption{DEPREL error comparison between PKT-UD v2018 and v2020.} 
\label{tbl:deprel-error-comparison}
\vspace{-2ex}
\end{table}


\section{Conclusion}
\label{sec:conclusion}

In this study, we revise the Penn Korean Universal Dependency Treebank (PKT-UD) and compare parsing performance between models trained on the original and revised versions of PKT.
Our new guidelines follow the UDv2 guidelines.
UPOS and DEPREL are revised to reflect Korean morphological features and flexible word-order aspects with reference to Korean UD studies such as \citet{park2018}, \citet{Lee2019}, and \citet{Oh2019}. In UPOS, \texttt{ADJ}, \texttt{NOUN}, and \texttt{VERB} are revised extensively. 
In DEPREL, \texttt{iobj}, \texttt{acl}, \texttt{advcl}, and \texttt{goeswith} are thoroughly revised. 
The revision results showing the percentage change of each label are presented in  Table~\ref{tbl:upos-stats} and Table~\ref{tbl:udep-stats}.

As a result of the parsing experiment, the v2020 model improves UAS by 0.6\% and LAS by 3.0\% over the v2018 model. 
In particular, \texttt{obl}, \texttt{acl}, \texttt{nummod}, and \texttt{advcl} errors are significantly reduced. 
This study, which improves parsing accuracy by applying characteristics of Korean, can also contribute to improve the quality of other Korean UD treebanks. 
In the future, we will explore the possibility of extending PKT-UD with enhanced dependency types\footnote{\url{https://universaldependencies.org/u/overview/enhanced-syntax.html}} by incorporating empty categories from the original PKT.

\bibliographystyle{acl_natbib}
\bibliography{iwpt2020}

\end{document}